\documentclass[11pt,a4paper]{article}
\usepackage[hyperref]{naaclhlt2018}
\usepackage{times}
\usepackage{latexsym}

\usepackage{url}

\aclfinalcopy 
\def\aclpaperid{117} 

\usepackage{times}
\usepackage{latexsym}
\usepackage{algorithm}
\usepackage[noend]{algpseudocode}
\usepackage{verbatim}
\usepackage{graphicx}
\usepackage{amsmath}
\usepackage{dblfloatfix}

\newcommand{\note}[1]{\textcolor{blue}{[#1]}}

\newcommand{\andy}[1]{\textcolor{red}{$_{A}$[#1]}}

\title{Predicting Human Trustfulness from Facebook Language}
 \author{Mohammadzaman Zamani \\ Computer Science Department \\ Stony Brook University \\ mzamani@cs.stonybrook.edu 
 \And Anneke Buffone \\Department of Psychology \\ University of Pennsylvania\\ buffonea@sas.upenn.edu  
 \And
 H. Andrew Schwartz \\ Computer Science Department \\ Stony Brook University \\ has@cs.stonybrook.edu }
 
 \date{}

\begin{document}
 \maketitle




\begin{abstract}
Trustfulness  --- one's general tendency to have confidence in unknown people or situations --- predicts many important real-world outcomes such as mental health and likelihood to cooperate with others such as clinicians. 
While data-driven measures of \textit{interpersonal trust} have previously been introduced, here, we develop the first language-based assessment of the personality trait of \textit{trustfulness} by fitting one's language to an accepted questionnaire-based trust score.
Further, using trustfulness as a type of case study, we explore the role of questionnaire size as well as word count in developing language-based predictive models of users' psychological traits.
We find that leveraging a longer questionnaire can yield greater test set accuracy, while, for training, we find it beneficial to include users who took smaller questionnaires which offers  
more observations for training. 
Similarly, after noting a decrease in individual prediction error as word count increased, we found a word count-weighted training scheme was helpful when there were very few users in the first place. 

\end{abstract}

\section{Introduction}




Trust, in general, indicates confidence that an entity or entities will behave in an expected manner~\cite{singh2007privacy}. 
While trust has been computationally explored as a property of relationships between people, i.e. \textit{interpersonal trust}
~\cite{golbeck2003trust,colquitt2007trust,murray2012resilience}, few have considered \textit{trustfulness} -- a personality trait of an individual indicating their tendency, outside of any other context, to trust in people, institutions, and situations~\cite{nannestad2008have}. 

\textit{Trustfulness} is tied to many real world and social outcomes. 
For example, it predicts individual  health~\cite{helliwell2010trust}, and how likely one is to join or to cooperate in diverse social groups~\cite{uslaner2002moral,stolle2002trusting}, and individual mental health and well-being~\cite{helliwell2010trust}.
The importance of trustfulness is thought to be increasing as modern societies are increasingly interacting online with unknown people~\cite{dinesen2016foundations}.
This suggests it could be increasingly important in a clinical domain where has been shown to be essential in securing a strong and effective patient-client bond~\cite{brennan2013trust, lambert2001research}. 
Trait trustfulness also relates to self-disclosure which in turn greatly aids the clinician in her provision of care~\cite{steel1991interpersonal}. Provider trust also likely is important to effectively treat a patient, especially in online therapeutic sessions, as it signals trustworthiness and care, but research on this topic remains sparse. 

Unfortunately, traditional trustfulness measurement options (e.g. surveys)  are expensive to scale to large populations and repeated assessment (i.e. in clinical practice) and they carry biases~\cite{baumeister2007psychology,youyou2017birds}. 
Researchers are actively searching for alternative behavior-based methods of measurement~\cite{nannestad2008have}. 


Language use in social media offers a behavior from which one can measure psychological traits like trust. 
Over the last five years, more and more researchers are turning to Facebook or Twitter language to develop psychological trait predictors, fitting user language to psychological scores from questionnaires~\cite{schwartz2015data}. 
According to standard psychometric validity tests, such language-based approaches have been found to rival other accepted measures, such as questionnaires and assessments from friends~\cite{park2015automatic}.
However, while language-based predictive models for many traits now exist, none have considered a model for trustfulness--- a trait which some have argued is now of marked importance as modern societies are increasingly interacting online with unknown people~\cite{dinesen2016foundations}.
Further, across such trait prediction work, little attention has been paid to the role of (1) \textit{questionnaire-size} -- how many questions are used to assess an individual's trait, and (2) \textit{word count} -- how many words the user has written from which the language-based predictions are made.\footnote{One often imposes a word count limitation --- e.g. users must write at least 1,000 words~\cite{schwartz2013personality}  --- but few have studied the relationship between word count and accuracy as we do here.}

Here, we answer the call for more behavior-based trait measurement~\cite{baumeister2007psychology,youyou2017birds}, by developing language-based (a behavior) predictive model of trustfulness fit to questionnaire scores, and we seek to draw insights into the role of word count and questionnaire size in predictive modeling.

\textbf{Contributions.} This work makes several key contributions. 
First, we introduce the first language-based assessment of \textit{trustfulness} (henceforth ``trust''), evaluated over out-of-sample trust questionnaires, enabling large-scale or frequently repeated trust measurement.
We also (2) study the number of questions in the psychological survey to which one fits our model (in other words, finding which one matters more: number of questions in questionnaires or number of users who took it?), (3) explore the relationship between users' word count and model error, and (4) introduce a weighting scheme to train on low word count users.
All together, we add trustfulness, an important trait for clinical care, to an increasing battery of language-based assessments.

\section{Background}





Previous computational work on trust has focused on \textit{interpersonal trust} -- an expectation of trust concerning future behaviour of a specific person toward another known person. \cite{bamberger2010interpersonal}. 
Interpersonal trust is primarily focused on situations in which there are two known individuals (the truster and trustee) who share a history of previous interactions. Such trust, requires study of a history of interactions indicating how well each member participant might understand the others' personalities~\cite{kelton2008trust,golbeck2003trust}.
Interpersonal trust has been studied especially in the context of online social networks where it is sometimes possible to track users from first interactions~\cite{kuter2007sunny,dubois2011predicting,liu2014assessment,liu2008predicting}. 
While some of these works have considered the amount of communication~\cite{adali2010measuring}, content is rarely considered and none of these past works have attempted to measure the trait, \textit{trustfulness}, as we do here. 

\textit{Trustfulness} (also referred to as ``generalized trust''), in contrast with interpersonal trust, measures trust between strangers. 
As ~\newcite{stolle2002trusting} put it:
\begin{quote}
\textit{[Trustfulness] indicates the potential readiness of citizens to cooperate with each other and to abstract preparedness to engage in civic endeavors with each other. Attitudes of trustfulness extend beyond the boundaries of face-to-face interactions and incorporate people who are not personally known.}
\end{quote}
This version of trust has been tied to the belief in the average goodness of human nature~\cite{yamagishi1994trust},  and it involves a willingness to be vulnerable and engage with random others despite interpersonal risks~\cite{mayer1999effect,rousseau1998not}. 
It has been shown predictive of individual mental health and physical well-being~\cite{abbott2008social,helliwell2010trust}.
For communities, trust is a key indicator of social capital~\cite{coleman1988social,putnam1993prosperous}, and it is highly predictive of economic growth~\cite{delhey2005predicting,knack2003building}

\subsection{Trustfulness from Questionnaires}
Trustfulness, just like other personality traits is typically measured with either questionnaires or behavioral observations during experiments~\cite{ermisch2009measuring}. 
Data linking experiments on trust with individual linguistic data is not available or easily acquired, so we fit our langauge-based model of trust to a gold-standard of questionnaire-based trust. 
A variety of such questionnaires exist with high inter-correlation, including the Faith in People scale (Rosenberg, 1957)\nocite{rosenberg1957occupation}, Yamagishi \& Yamagishi’s (1999) Trust Scale\nocite{yamagishi1999trust}, and the Trust Facet of the Agreeableness trait in the Big Five personality questionnaire~\cite{goldberg2006international}. 
Here, due to its availability, we chose to fit our language-based trust predictor to the later of these questionnaires -- the trust personality facet.


\section{Data Set.}
We use trust facet scores from the trait questionnaire of consenting participants of the MyPersonality study~\cite{kosinski2015facebook}.\footnote{Procedures were approved by an academic institutional review board, and volunteers agreed to share their data for research purposes via informed consent.} From this dataset we derive two versions of trust measurement scores: (1) using $10$ questions of trustfulness (referred to as \textit{10-question trust}), or (2) using a subset of $3$ questions (referred to as \textit{3-question trust}). 
Participants can either answer all 10 questions (as part of larger set of over 300 questions) or just answer the 3-question version (as part of a 100 questions). 
Each question is on a scale of 1 to 5, from totally disagree to completely agree. 
For example, the following are the questions for the 3-item version: 
\begin{itemize}
\setlength{\itemsep}{0pt}
    \item \textit{I believe that others have good intentions.}
    \item \textit{I suspect hidden motives in others.$^*$} 
    \item \textit{I trust what people say. }
\end{itemize}
Some questions (e.g. $^*$ above) are ``reverse scored'' so a 1 becomes a 5 and vice-versa. 
One's final trust score is based on taking the mean of the responses to the individual trust questions. 
Although 3-question trust is less accurate,\footnote{
Based on an experiment across $1,041$ participants on Amazon Mechanical Turk, 3-question trust had a Pearson correlation of $0.916$ with 10-question trust, indicating it is a reasonable approximation~\cite{Buffone2017Measuring}} it may be useful to enable training data from more users.



%



From MyPersonality, we used a dataset containing $19,455$ Facebook users who wrote at least $1,000$ words across all of their status updates. We additionally included $6,590$ users who had less than $1,000$ words in some experiments. Totally $26,045$ users took the Big Five questionnaire, answering at least the 3 trust-focused questions in it (short version). Among all the users, only $621$ had completely answered all of the 10 trust related question (long version). Table~\ref{table:users} represents number of users in detail. It is worth mentioning that not only the participants consent for their Facebook and questionnaire data to be used in research, but also the data has been anonymized.

\begin{table}[ht]
\centering
\label{table:users}
\begin{tabular}{|l||c|c|}
\hline
               & Long Version & Short Version \\
\hline
\hline
Threshold-1000 & 438          & 19445         \\
\hline
Threshold-10   & 621          & 26045       \\ 
\hline
\end{tabular}
\caption{Number of users who filled the long or short version questionnaire based on their word counts. Threshold-X means setting word count threshold to X. Long version represents users who had 10-question trust score, and short version includes users who had 3-question trust score.  }
\end{table}

\section { Method }
We build a language-based model for the trait of trustfulness. 
From Facebook status updates, we extracted two types of user-level lexical features, which have previously been shown to be effective for trait prediction \cite{park2015automatic}: (a) ngrams of length 1 to 3 and (b) LDA topics. To extract the ngrams from the text we used the \textit{HappierFunTokenizer}. We did not apply any text normalization, as past work has found that often the forms in which people choose to write a word ends up being predictive about their personality~\cite{schwartz2013personality}.
Two types of ngrams were extracted: one containing relative frequencies of each ngram ($\frac{freq(ngram,user)}{freq(*,user)}$) and the other simply a binary indicator of whether the user mentioned each ngram at all. Considering ngrams mentioned by at least $1\%$ of the users, we obtained $50,166$ ngrams features for each of the two types of ngrams.
Topic features were derived from posteriors of Latent Dirichlet Allocation.  We use the $2,000$ LDA topic posteriors publicly available from Schwartz et al. (2013).\footnote{Topic posteriors and the tokenizer, Happier Fun Tokenizing, are available from http://wwbp.org/data.html}. 


We use a series of steps to avoid high dimensional issues and prevent overfitting.
%
%
First, an occurrence threshold is applied to remove words that were used by less than 1\% of people. 
Second, we select features with at least a small relationship with our trust labels according to having a univariate family-wise error rate $< 60$. 
Third, we ran a singular value decomposition (in randomized batches) to effectively decrease the size of feature space and reduce colinearity across dimensions\cite{boutsidis2015randomized}. %
We performed this process based on the training data, and then applied the resulting feature reduction on the test data.

Each type of feature (i.e. ngram relative frequencies, booleans, and topics) is qualitatively and distributionally different from each other~\cite{almodaresi2017distribution}. 
Thus, we perform reduction technique on ngrams, boolean ngrams and topics separately. 
This is so the comparatively few topic features are not likely to get lost among the relatively plentiful ngrams.
At the end, we merge both types of features to build a single feature matrix (or an embedding with approximately 5\% of the number of training observations). 
Similar feature reduction pipelines have been shown to perform well in language based predictive analytics~\cite{zamani2017using}.
We then use ridge regression to fit our dimensionally reduced feature set to the trust labels from the Big Five questionnaires. 
%

\paragraph*{Questionnaire Size and Word Count.}
While the 10-question trust score is more accurate, we  have less than $1,000$ users with this label.
Our default setup has the users with 10-question trust as the test set while we train over the much larger set of users with only 3-question trust. 
We then experiment to determine if this setup is ideal. 

Previous work has suggested user attribute prediction benefits from an approximate minimum threshold of $1,000$ words per user in order to get accurate estimates of one's personality~\cite{schwartz2013personality}. 
Since our dataset contains $6,590$ users with less than $1,000$ words, we explore if we can include these users in an effective way to improve the model. 
To this end, we weight each users' contribution to the loss function proportionate to the number of words she or he has written. 
We used two different weighting schemes, linear and logistic, as shown below, where $wc$ is the word count, and $T_{max}$ and $T_{min}$ are $1,000$ and $200$ respectively, and $\alpha$ is a parameter that should be a reasonably big number as we selected to be 100.
$$   
W_{linear} = \frac{\min(T_{max}, max(T_{min}, {wc}) ) -T_{min}}{T_{max} - T_{min}}
$$
$$
 W_{logistic} = \frac{1}{1+ \exp( -\alpha * (W_{linear} - 1/2) )}
$$

Thus, users with more than $1,000$ words are weighted $1$ while those with less than $200$ words are weighted $0$ (we settled on these min and max values based on our study of the mean error per word count -- Figure \ref{fig:uwt_error}).

\section{Evaluation}

We focus on evaluating our language model by comparing the performance of our model on prediction of 10-question trust vs. 3-question trust labels. We did this comparison in 3 settings: (1) train and test on 10-question trust score, (2) train and test on 3-question trust score, and (3) train on 3-question and test on 10-question trust score.

For the first setting, where all users answered the same number of questions, we performed a 10-fold cross-validation. 
For the second and third settings, we consider all users with 10-question trust score as our test group and the remaining users which only had 3-question trust score but not the 10-question trust as the train group. 
This enables us to first determine how well a model trained on 3-question trust performs in not only predicting 3-question trust itself, but also the 10-question trust, and compare the later with the model which is trained on small group of users with 10-question trust. 
In all these three experiments, we considered $1,000$ as the threshold for word count, and used the same group of users as the test group.
We present result as both \textit{mean squared error} and \textit{disattenuated correlation} which accounts for measurement error: $r_{dis(a,b)} = \frac{r_{a,b}}{\sqrt{r_{a,a}r_{b,b}}}$ where $r_{a,a} = .70$ the reliability of the trust questionnaire~\cite{kosinski2015facebook} and $r_{b,b} = .70$ the expected reliability of the trust language-based measurement based on evaluations of language-based personality assessment reliability~\cite{park2015automatic} (every $r$ on the right-hand side of the equation is a Pearson product-moment correlation coefficient). 
%



\begin{table}[th]
\centering
\small
\begin{tabular}{|l|c|c|c|c|}
\hline
  train label          & test label & \textbf{Pearson $r_{dis}$} & \textbf{MSE}  \\ \hline
10-question & 10-question & 0.259 & 0.719  \\ \hline
3-question  & 3-question  & 0.426              & 0.776               \\ \hline
3-question & 10-question  &\textbf{0.494} & \textbf{0.662}    \\ \hline
\end{tabular}
\caption {Comparing the language model performance on 3-question trust score vs. 10-question trust score. Pearson $r_{dis}$ is dissattenuated Pearson r and MSE is the mean squared error. }
\label{table:3vs10}
\end{table}

As shown in table \ref{table:3vs10}, our model's $r_{dis}$ with only limited 10-item data is $0.259$, suggesting we cannot learn a very accurate model by training on such a small number of users. 
Comparing the second and third settings, we see the result of testing on 10-question trust score outperforms the 3-question trust score by $0.07$ margin in dissattenuated Pearson r and MSE by a margin of $0.11$. 
To further understand why 10-question trust seems to be easier to predict, we calculate the variance for both 3-question and 10-question trust, yielding $\sigma^2 = 0.85$ and $\sigma^2 = 0.72$ respectively. 
This suggests that 10-question trust has less noise than 3-question trust. 
Due to these results, in all of the following experiments we only train on 3-question trust labels and test on 10-question trust labels.

We next evaluate the performance of our trust model by comparing to two baseline models.
Because positiveness is associated with trust~\cite{helliwell2010trust}, we consider a baseline of sentiment scores using the NRC hashtag sentiment lexicon, an integral part of the best system participating in SemEval-2013~\cite{mohammad2013nrc}.
We also compare it to clusters of words derived from word2vec embeddings~\cite{mikolov2013efficient} using spectral clustering~\cite{preoctiuc2015analysis}.  

\begin{table}[t]
\centering
\small
\begin{tabular}{|l|c|c|}
\hline
\textbf{Features}                & \textbf{Pearson $r_{dis}$} & \textbf{MSE}   \\ \hline
sentiment (baseline)               & 0.279     & 0.717 \\ \hline
ngr\_r                   & 0.453     & 0.681 \\ \hline
ngr\_b                  & 0.411     & 0.688 \\ \hline
topics                   & 0.458     & 0.677  \\ \hline
word2vec & 0.449 &   0.678   \\ \hline 
ngr\_r + ngr\_b +topics      & \textbf{0.494}     & \textbf{0.662} \\ \hline
ngr\_r + ngr\_b +topics+sent & 0.483     & 0.666 \\ \hline
\end{tabular}
\caption{Comparing the performance of our language model with sentiment as baseline, using different feature sets: ngr\_r: ngrams as relative frequencie, ngr\_b: ngrams as boolean variables. Bold indicates the best performance. Pearson $r_{dis}$ is dissattenuated Pearson r and MSE is the mean squared error.}
\label{table:separated_feats}
\end{table}

\begin{figure}[ht]
    \centering
    \includegraphics[width=200px]{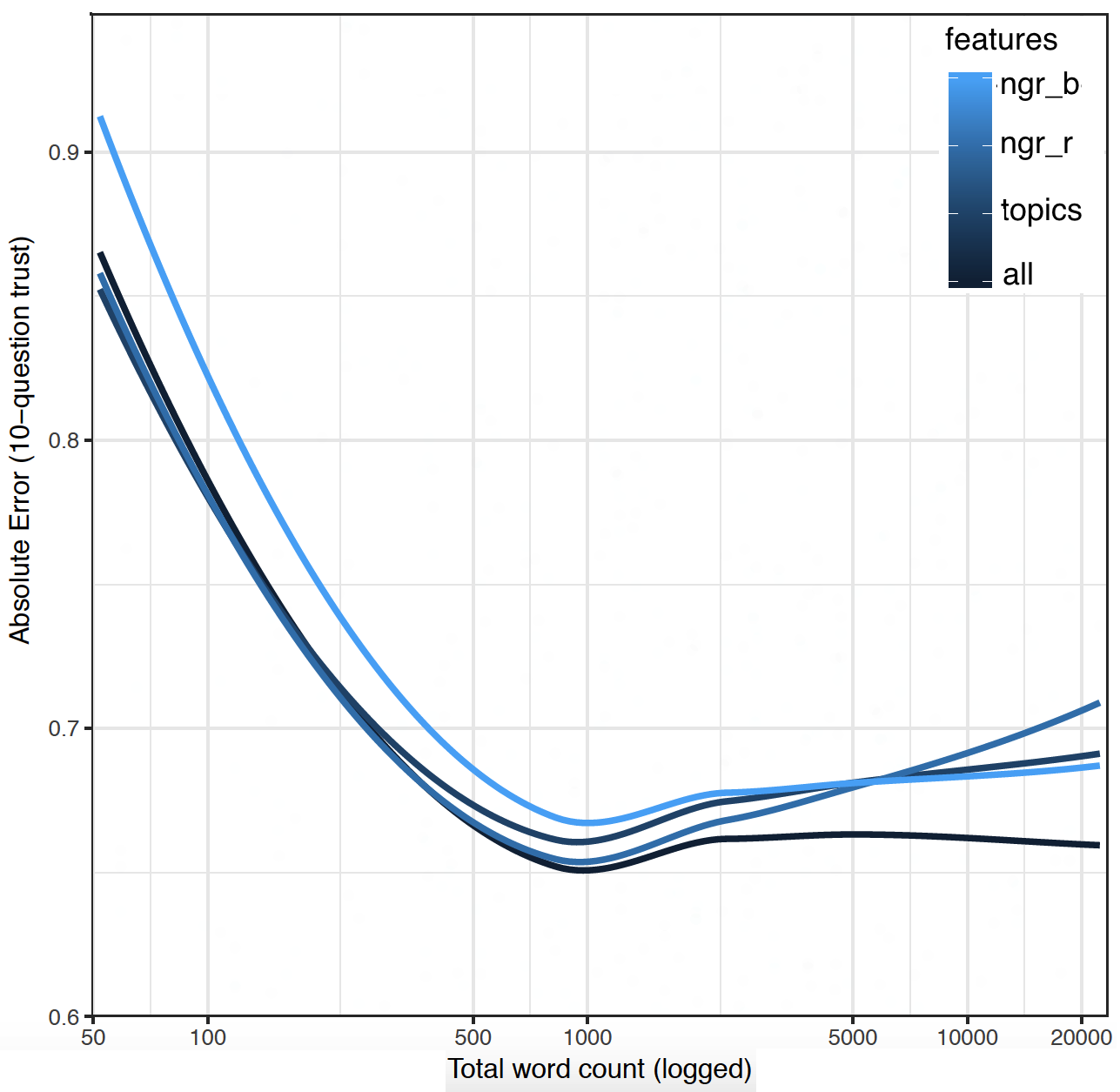}
    \caption{Effect of word count on error rate of the language model: ngr\_b stands for binary-ngrams and ngr\_r stands for relative-ngrams.}
    \label{fig:uwt_error}
\end{figure}

Table~\ref{table:separated_feats} demonstrates the predictive performance of our model in comparison to the sentiment and word2vec baselines. 
Our best model ($ngr\_r + ngr\_b +topics$) had an $8\%$ reduction in mean squared error over sentiment, and achieved a Pearson correlation coefficient of $r_{dis} = .494$ which is considered a large relationship between a behavior (language use) and a psychological trait~\cite{meyer2001psychological} and just below state-of-the-art language-based assessments of other personality traits~\cite{park2015automatic}.

In the next experiment we present how the error rate changes as a function of word count per user using various combinations of features.
We trained 4 models using (1) relative-ngrams, (2) binary-ngrams, (3) topics, and (4) all features together. 
We predict the 10-question trust score of our test users and plot the test users error rate with respect to their word count, which is shown in figure~\ref{fig:uwt_error}. 
Overall, users' trust score is more predictable as they use more words flattening out after 1000 words.
Additionally, for users with few words, relative-ngrams and binary-ngrams are equally predictive and better than topics. For users with many words, the prediction power of binary-ngrams fades out, likely reflecting features being primarily ones. 
Similarly, topic-based models perform better for talkative users, likely because more words means better topic estimation.

\begin{figure}[ht]
\includegraphics[width=225pt,height=180pt]{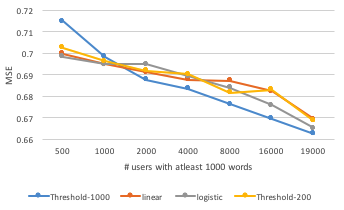}
\caption{Effect of increasing the number of training users, who have more than $1,000$ word count, while there are $6,590$ users with less than $1,000$ word count in train set: ``Threshold-1000'' is training ridge-regression on users with at least $1,000$ words, ``threshold-200'' is training ridge-regression on users with at least $200$ words, ``linear'' is training weighted ridge-regression on users with at least $200$ words, and finally ``logistic'' is training weighted ridge-regression on users with at least $200$ words.}
\label{fig:weighting}
\end{figure}

\begin{figure*}[ht]
\centering
    \includegraphics[width=.36\textwidth]{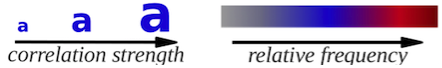}\\
  \begin{tabular}{cc}
    \includegraphics[width=.5\textwidth]{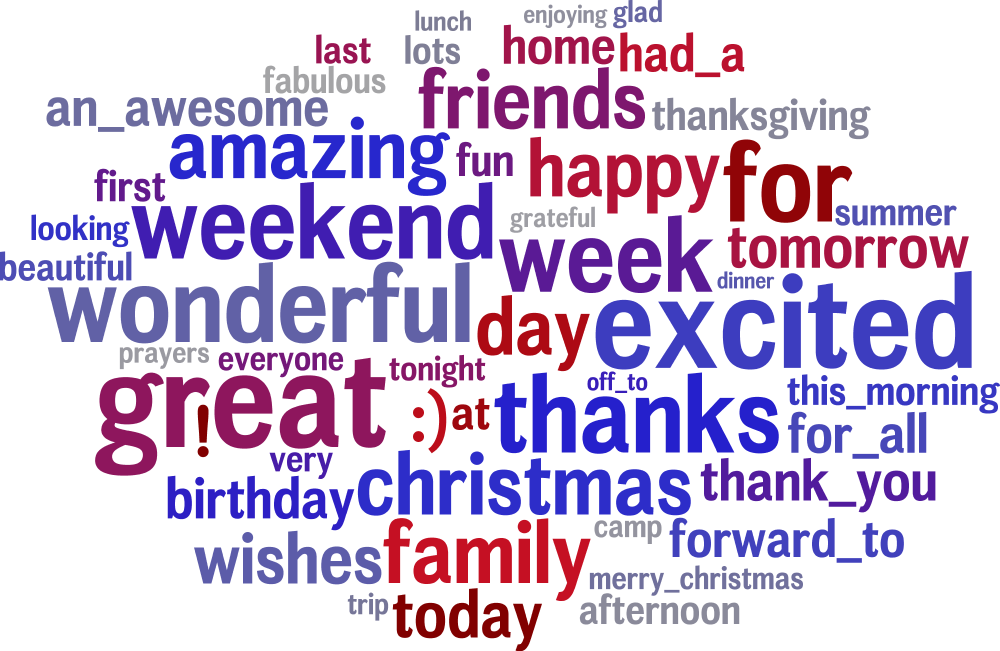} &
    \includegraphics[width=.45\textwidth]{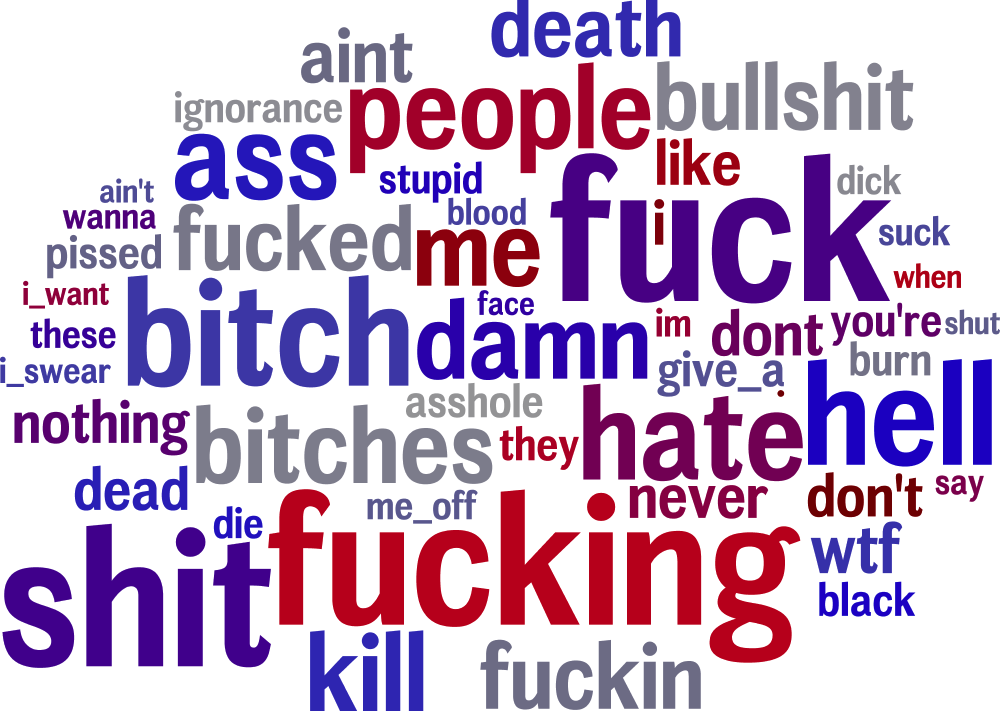}   \\
    \includegraphics[width=.25\textwidth]{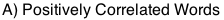} &
    \includegraphics[width=.25\textwidth]{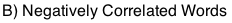}   \\
  \end{tabular}
  \caption{Unigrams most distinguish trust according to absolute value of (a) positive correlation and (b) negative correlated with 3-question trust score. Size of word indicates correlation strength, while color indicates frequency. All unigrams listed are significantly correlated at Benjamini-Hochberg corrected $p < .05$.}
  \label{fig:pos_neg_cloud}
\end{figure*}

Now that we know  word count is correlated with prediction error, we explore a word count weighting scheme that enables us to include $6,590$ users with fewer than $1,000$ words in training. 
Such users are included in three different ways, (1) without using any weight, (2) using linear weighting, and (3) using logistic weighting. 

In figure~\ref{fig:weighting} we compare the various model training setups at different training sizes. 
As shown, when we have just a few users with more than $1,000$ words, including more users, but with low word count, improves the performance, no matter which models we exploit. 
However, as the number of users with more than $1,000$ word count increases, injecting low word count users hurts the performance. 
In addition, the weighting scheme does not seem to help at all in this situation. 

To get an idea of the type of features signalling high and low trust predictions, we ran a differential language analysis~\cite{schwartz2013personality} to identify the top 50, independently, most predictive features.
Figure~\ref{fig:pos_neg_cloud} show the word-clouds of both positively correlated and negatively correlated with 3-question trust score, limited to those passing a Benjamini-Hochberg False Discovery rate $alpha$ of $0.01$~\cite{benjamini1995controlling}. 
Many of the ngrams correspond with the definition of trustfulness, such as the pro-social words in the positive predictors (e.g. `friends' `family', `thanks'). On the other hand, many curse words can be seen among negative predictors.  

\section{Conclusion}

We introduced the first language-based model for measuring \textit{trustfulness} from language, and used it to study novel and useful aspects of the predictive modeling of user traits. 
First, we found that language use in social media can be used to predict trustfulness about as accurate as other personality traits. 
Then, we found that, in order to build a language model over questionnaires, including \textit{more} users who took a \textit{shorter} questionnaire can lead to improvement, in comparison to using \textit{less} users who took a \textit{longer} questionnaire.
We also showed that the language model usually performs better in predicting users with more total word count, with error flattening out around $1,000$ words,  
and that when there are few users (i.e $< 1,000$) it is worth lowering the minimum word count threshold to include more users for training purpose.
However, using a weighting scheme was not helpful. 

Our scaleable measure of trust enables future work to investigate some interesting questions about trust, such as those involved in large-scale or frequent assessments. For example, this may allow for large-scale assessments of trait trustfulness of different patient populations or of samples of clinicians. Also, if clients were to opt into sharing of social media, therapists may be able to use this model to detect drops in patient trust which may help to understand when one is more receptive or not. Trends over time may help to signal interpersonal improvements or regressions, as well as negative interactions with others. It should be noted that while trust is thought of as a relatively stable personality aspect or trait, some research suggests that it is malleable over time~\cite{jones1998experience}, so changes in trust over time could be another meaningful exploration for future study. Thus, the present model may be helpful for the generation of trustful chat bots, such as virtual assistants or therapeutic aids. 
\bibliographystyle{acl_natbib}

\begin{thebibliography}{47}
\expandafter\ifx\csname natexlab\endcsname\relax\def\natexlab#1{#1}\fi

\bibitem[{Abbott and Freeth(2008)}]{abbott2008social}
Stephen Abbott and Della Freeth. 2008.
\newblock Social capital and health: starting to make sense of the role of
  generalized trust and reciprocity.
\newblock \emph{Journal of Health Psychology}, 13(7):874--883.

\bibitem[{Adali et~al.(2010)Adali, Escriva, Goldberg, Hayvanovych,
  Magdon-Ismail, Szymanski, Wallace, Williams et~al.}]{adali2010measuring}
Sibel Adali, Robert Escriva, Mark~K Goldberg, Mykola Hayvanovych, Malik
  Magdon-Ismail, Boleslaw~K Szymanski, William Wallace, Gregory Williams,
  et~al. 2010.
\newblock Measuring behavioral trust in social networks.
\newblock In \emph{Intelligence and Security Informatics (ISI), 2010 IEEE
  International Conference on}, pages 150--152. IEEE.

\bibitem[{Almodaresi et~al.(2017)Almodaresi, Ungar, Kulkarni, Zakeri, Giorgi,
  and Schwartz}]{almodaresi2017distribution}
Fatemeh Almodaresi, Lyle Ungar, Vivek Kulkarni, Mohsen Zakeri, Salvatore
  Giorgi, and H~Andrew Schwartz. 2017.
\newblock On the distribution of lexical features at multiple levels of
  analysis.
\newblock In \emph{Proceedings of the 55th Annual Meeting of the Association
  for Computational Linguistics}, volume~2, pages 79--84.

\bibitem[{Bamberger(2010)}]{bamberger2010interpersonal}
Walter Bamberger. 2010.
\newblock Interpersonal trust--attempt of a definition.
\newblock \emph{Scientific Report, Technical University Munich}.

\bibitem[{Baumeister et~al.(2007)Baumeister, Vohs, and
  Funder}]{baumeister2007psychology}
Roy~F Baumeister, Kathleen~D Vohs, and David~C Funder. 2007.
\newblock Psychology as the science of self-reports and finger movements:
  Whatever happened to actual behavior?
\newblock \emph{Perspectives on Psychological Science}, 2(4):396--403.

\bibitem[{Benjamini and Hochberg(1995)}]{benjamini1995controlling}
Yoav Benjamini and Yosef Hochberg. 1995.
\newblock Controlling the false discovery rate: a practical and powerful
  approach to multiple testing.
\newblock \emph{Journal of the royal statistical society. Series B
  (Methodological)}, pages 289--300.

\bibitem[{Boutsidis et~al.(2015)Boutsidis, Zouzias, Mahoney, and
  Drineas}]{boutsidis2015randomized}
Christos Boutsidis, Anastasios Zouzias, Michael~W Mahoney, and Petros Drineas.
  2015.
\newblock Randomized dimensionality reduction for $ k $-means clustering.
\newblock \emph{IEEE Transactions on Information Theory}, 61(2):1045--1062.

\bibitem[{Brennan et~al.(2013)Brennan, Barnes, Calnan, Corrigan, Dieppe, and
  Entwistle}]{brennan2013trust}
Nicola Brennan, Rebecca Barnes, Mike Calnan, Oonagh Corrigan, Paul Dieppe, and
  Vikki Entwistle. 2013.
\newblock Trust in the health-care provider--patient relationship: a systematic
  mapping review of the evidence base.
\newblock \emph{International Journal for Quality in Health Care},
  25(6):682--688.

\bibitem[{Buffone et~al.(2017)Buffone, Schwartz, Crutchley, Kern, Zamani,
  Smith, Eichstaedt, Ungar, and M.~Seligman}]{Buffone2017Measuring}
Anneke Buffone, H~Andrew Schwartz, Patrick Crutchley, Margaret~L. Kern,
  Mohammadzaman Zamani, L.~K. Smith, Johannes~C. Eichstaedt, Lyle Ungar, and
  Martin E.~P. M.~Seligman. 2017.
\newblock Measuring trust through large scale language analysis: Trust as an
  aspect of individuals and communities.
\newblock \emph{In Press}.

\bibitem[{Coleman(1988)}]{coleman1988social}
James~S Coleman. 1988.
\newblock Social capital in the creation of human capital.
\newblock \emph{American journal of sociology}, 94:S95--S120.

\bibitem[{Colquitt et~al.(2007)Colquitt, Scott, and LePine}]{colquitt2007trust}
Jason~A Colquitt, Brent~A Scott, and Jeffery~A LePine. 2007.
\newblock Trust, trustworthiness, and trust propensity: a meta-analytic test of
  their unique relationships with risk taking and job performance.
\newblock \emph{Journal of applied psychology}, 92(4):909.

\bibitem[{Delhey and Newton(2005)}]{delhey2005predicting}
Jan Delhey and Kenneth Newton. 2005.
\newblock Predicting cross-national levels of social trust: global pattern or
  nordic exceptionalism?
\newblock \emph{European Sociological Review}, 21(4):311--327.

\bibitem[{Dinesen and Bekkers(2016)}]{dinesen2016foundations}
Peter~Thisted Dinesen and Rene Bekkers. 2016.
\newblock The foundations of individuals’ generalized social trust: A review.
\newblock In \emph{Trust in Social Dilemmas}. Oxford University Press.

\bibitem[{DuBois et~al.(2011)DuBois, Golbeck, and
  Srinivasan}]{dubois2011predicting}
Thomas DuBois, Jennifer Golbeck, and Aravind Srinivasan. 2011.
\newblock Predicting trust and distrust in social networks.
\newblock In \emph{Privacy, Security, Risk and Trust (PASSAT) and 2011 IEEE
  Third Inernational Conference on Social Computing (SocialCom), 2011 IEEE
  Third International Conference on}, pages 418--424. IEEE.

\bibitem[{Ermisch et~al.(2009)Ermisch, Gambetta, Laurie, Siedler, and
  Noah~Uhrig}]{ermisch2009measuring}
John Ermisch, Diego Gambetta, Heather Laurie, Thomas Siedler, and
  SC~Noah~Uhrig. 2009.
\newblock Measuring people's trust.
\newblock \emph{Journal of the Royal Statistical Society: Series A (Statistics
  in Society)}, 172(4):749--769.

\bibitem[{Golbeck et~al.(2003)Golbeck, Parsia, and Hendler}]{golbeck2003trust}
Jennifer Golbeck, Bijan Parsia, and James Hendler. 2003.
\newblock \emph{Trust networks on the semantic web}.
\newblock Springer.

\bibitem[{Goldberg et~al.(2006)Goldberg, Johnson, Eber, Hogan, Ashton,
  Cloninger, and Gough}]{goldberg2006international}
Lewis~R Goldberg, John~A Johnson, Herbert~W Eber, Robert Hogan, Michael~C
  Ashton, C~Robert Cloninger, and Harrison~G Gough. 2006.
\newblock The international personality item pool and the future of
  public-domain personality measures.
\newblock \emph{Journal of Research in personality}, 40(1):84--96.

\bibitem[{Helliwell and Wang(2010)}]{helliwell2010trust}
John~F Helliwell and Shun Wang. 2010.
\newblock Trust and well-being.
\newblock Technical report, National Bureau of Economic Research.

\bibitem[{Jones and George(1998)}]{jones1998experience}
Gareth~R Jones and Jennifer~M George. 1998.
\newblock The experience and evolution of trust: Implications for cooperation
  and teamwork.
\newblock \emph{Academy of management review}, 23(3):531--546.

\bibitem[{Kelton et~al.(2008)Kelton, Fleischmann, and
  Wallace}]{kelton2008trust}
Kari Kelton, Kenneth~R Fleischmann, and William~A Wallace. 2008.
\newblock Trust in digital information.
\newblock \emph{Journal of the American Society for Information Science and
  Technology}, 59(3):363--374.

\bibitem[{Knack and Zak(2003)}]{knack2003building}
Stephen Knack and Paul~J Zak. 2003.
\newblock Building trust: public policy, interpersonal trust, and economic
  development.
\newblock \emph{Supreme court economic review}, 10:91--107.

\bibitem[{Kosinski et~al.(2015)Kosinski, Matz, Gosling, Popov, and
  Stillwell}]{kosinski2015facebook}
Michal Kosinski, Sandra~C Matz, Samuel~D Gosling, Vesselin Popov, and David
  Stillwell. 2015.
\newblock Facebook as a research tool for the social sciences: Opportunities,
  challenges, ethical considerations, and practical guidelines.
\newblock \emph{American Psychologist}, 70(6):543.

\bibitem[{Kuter and Golbeck(2007)}]{kuter2007sunny}
Ugur Kuter and Jennifer Golbeck. 2007.
\newblock Sunny: A new algorithm for trust inference in social networks using
  probabilistic confidence models.
\newblock In \emph{AAAI}, volume~7, pages 1377--1382.

\bibitem[{Lambert and Barley(2001)}]{lambert2001research}
Michael~J Lambert and Dean~E Barley. 2001.
\newblock Research summary on the therapeutic relationship and psychotherapy
  outcome.
\newblock \emph{Psychotherapy: Theory, research, practice, training},
  38(4):357.

\bibitem[{Liu et~al.(2014)Liu, Yang, Wang, Lin, and Wittie}]{liu2014assessment}
Guangchi Liu, Qing Yang, Honggang Wang, Xiaodong Lin, and Mike~P Wittie. 2014.
\newblock Assessment of multi-hop interpersonal trust in social networks by
  three-valued subjective logic.
\newblock In \emph{INFOCOM, 2014 Proceedings IEEE}, pages 1698--1706. IEEE.

\bibitem[{Liu et~al.(2008)Liu, Lim, Lauw, Le, Sun, Srivastava, and
  Kim}]{liu2008predicting}
Haifeng Liu, Ee-Peng Lim, Hady~W Lauw, Minh-Tam Le, Aixin Sun, Jaideep
  Srivastava, and Young Kim. 2008.
\newblock Predicting trusts among users of online communities: an epinions case
  study.
\newblock In \emph{Proceedings of the 9th ACM Conference on Electronic
  Commerce}, pages 310--319. ACM.

\bibitem[{Mayer and Davis(1999)}]{mayer1999effect}
Roger~C Mayer and James~H Davis. 1999.
\newblock The effect of the performance appraisal system on trust for
  management: A field quasi-experiment.
\newblock \emph{Journal of applied psychology}, 84(1):123.

\bibitem[{Meyer et~al.(2001)Meyer, Finn, Eyde, Kay, Moreland, Dies, Eisman,
  Kubiszyn, and Reed}]{meyer2001psychological}
Gregory~J Meyer, Stephen~E Finn, Lorraine~D Eyde, Gary~G Kay, Kevin~L Moreland,
  Robert~R Dies, Elena~J Eisman, Tom~W Kubiszyn, and Geoffrey~M Reed. 2001.
\newblock Psychological testing and psychological assessment: A review of
  evidence and issues.
\newblock \emph{American psychologist}, 56(2):128.

\bibitem[{Mikolov et~al.(2013)Mikolov, Chen, Corrado, and
  Dean}]{mikolov2013efficient}
Tomas Mikolov, Kai Chen, Greg Corrado, and Jeffrey Dean. 2013.
\newblock Efficient estimation of word representations in vector space.
\newblock \emph{arXiv preprint arXiv:1301.3781}.

\bibitem[{Mohammad et~al.(2013)Mohammad, Kiritchenko, and
  Zhu}]{mohammad2013nrc}
Saif~M Mohammad, Svetlana Kiritchenko, and Xiaodan Zhu. 2013.
\newblock Nrc-canada: Building the state-of-the-art in sentiment analysis of
  tweets.
\newblock \emph{arXiv preprint arXiv:1308.6242}.

\bibitem[{Murray et~al.(2012)Murray, Lupien, and Seery}]{murray2012resilience}
Sandra~L Murray, Shannon~P Lupien, and Mark~D Seery. 2012.
\newblock Resilience in the face of romantic rejection: The automatic impulse
  to trust.
\newblock \emph{Journal of Experimental Social Psychology}, 48(4):845--854.

\bibitem[{Nannestad(2008)}]{nannestad2008have}
Peter Nannestad. 2008.
\newblock What have we learned about generalized trust, if anything?
\newblock \emph{Annu. Rev. Polit. Sci.}, 11:413--436.

\bibitem[{Park et~al.(2015)Park, Schwartz, Eichstaedt, Kern, Kosinski,
  Stillwell, Ungar, and Seligman}]{park2015automatic}
Gregory Park, H~Andrew Schwartz, Johannes~C Eichstaedt, Margaret~L Kern, Michal
  Kosinski, David~J Stillwell, Lyle~H Ungar, and Martin~EP Seligman. 2015.
\newblock Automatic personality assessment through social media language.
\newblock \emph{Journal of personality and social psychology}, 108(6):934.

\bibitem[{Preo{\c{t}}iuc-Pietro et~al.(2015)Preo{\c{t}}iuc-Pietro, Lampos, and
  Aletras}]{preoctiuc2015analysis}
Daniel Preo{\c{t}}iuc-Pietro, Vasileios Lampos, and Nikolaos Aletras. 2015.
\newblock An analysis of the user occupational class through twitter content.

\bibitem[{Putnam(1993)}]{putnam1993prosperous}
Robert~D Putnam. 1993.
\newblock The prosperous community.
\newblock \emph{The american prospect}, 4(13):35--42.

\bibitem[{Rosenberg(1957)}]{rosenberg1957occupation}
Morris Rosenberg. 1957.
\newblock Occupation and values: Glencoe.

\bibitem[{Rousseau et~al.(1998)Rousseau, Sitkin, Burt, and
  Camerer}]{rousseau1998not}
Denise~M Rousseau, Sim~B Sitkin, Ronald~S Burt, and Colin Camerer. 1998.
\newblock Not so different after all: A cross-discipline view of trust.
\newblock \emph{Academy of management review}, 23(3):393--404.

\bibitem[{Schwartz et~al.(2013)Schwartz, Eichstaedt, Kern, Dziurzynski,
  Ramones, Agrawal, Shah, Kosinski, Stillwell, Seligman
  et~al.}]{schwartz2013personality}
H~Andrew Schwartz, Johannes~C Eichstaedt, Margaret~L Kern, Lukasz Dziurzynski,
  Stephanie~M Ramones, Megha Agrawal, Achal Shah, Michal Kosinski, David
  Stillwell, Martin~EP Seligman, et~al. 2013.
\newblock Personality, gender, and age in the language of social media: The
  open-vocabulary approach.
\newblock \emph{PloS one}, 8(9):e73791.

\bibitem[{Schwartz and Ungar(2015)}]{schwartz2015data}
H~Andrew Schwartz and Lyle~H Ungar. 2015.
\newblock Data-driven content analysis of social media: a systematic overview
  of automated methods.
\newblock \emph{The ANNALS of the American Academy of Political and Social
  Science}, 659(1):78--94.

\bibitem[{Singh and Bawa(2007)}]{singh2007privacy}
Sarbjeet Singh and Seema Bawa. 2007.
\newblock A privacy, trust and policy based authorization framework for
  services in distributed environments.
\newblock \emph{International Journal of Computer Science}, 2(2):85--92.

\bibitem[{Steel(1991)}]{steel1991interpersonal}
Jennifer~L Steel. 1991.
\newblock Interpersonal correlates of trust and self-disclosure.
\newblock \emph{Psychological Reports}, 68(3\_suppl):1319--1320.

\bibitem[{Stolle(2002)}]{stolle2002trusting}
Dietlind Stolle. 2002.
\newblock Trusting strangers--the concept of generalized trust in perspective.
\newblock \emph{Austrian Journal of Political Science}, 31(4):397--412.

\bibitem[{Uslaner(2002)}]{uslaner2002moral}
Eric~M Uslaner. 2002.
\newblock \emph{The moral foundations of trust}.
\newblock Cambridge University Press.

\bibitem[{Yamagishi et~al.(1999)Yamagishi, Kikuchi, and
  Kosugi}]{yamagishi1999trust}
Toshio Yamagishi, Masako Kikuchi, and Motoko Kosugi. 1999.
\newblock Trust, gullibility, and social intelligence.
\newblock \emph{Asian Journal of Social Psychology}, 2(1):145--161.

\bibitem[{Yamagishi and Yamagishi(1994)}]{yamagishi1994trust}
Toshio Yamagishi and Midori Yamagishi. 1994.
\newblock Trust and commitment in the united states and japan.
\newblock \emph{Motivation and emotion}, 18(2):129--166.

\bibitem[{Youyou et~al.(2017)Youyou, Schwartz, Stillwell, and
  Kosinski}]{youyou2017birds}
Wu~Youyou, H~Andrew Schwartz, David Stillwell, and Michal Kosinski. 2017.
\newblock Birds of a feather do flock together: Behavior-based
  personality-assessment method reveals personality similarity among couples
  and friends.
\newblock \emph{Psychological Science}, page 0956797616678187.

\bibitem[{Zamani and Schwartz(2017)}]{zamani2017using}
Mohammadzaman Zamani and H~Andrew Schwartz. 2017.
\newblock Using twitter language to predict the real estate market.
\newblock In \emph{Proceedings of the 15th Conference of the European Chapter
  of the Association for Computational Linguistics: Volume 2, Short Papers},
  volume~2, pages 28--33.

\end{thebibliography}

\end{document}